\documentclass[runningheads]{llncs}
\usepackage[linesnumbered,boxed,ruled,commentsnumbered,vlined]{algorithm2e}
\usepackage{float}
\usepackage{graphicx}
\usepackage{wrapfig}
\usepackage{picinpar}
\usepackage{cutwin}
\usepackage{subfigure}
\usepackage{textcomp}
\usepackage{xcolor}
\usepackage{multirow}
\usepackage{lineno}
\usepackage{bbding}

\usepackage{moresize}

\usepackage[stable]{footmisc} 
\usepackage{amsmath}
\usepackage{amssymb}
\usepackage[utf8]{inputenc} 
\usepackage{array}
\usepackage{threeparttable} 
\usepackage{booktabs} 
\usepackage{hyperref}
\usepackage{arydshln}
\usepackage{subcaption}

\hypersetup{colorlinks=true,       
            linkcolor=blue,       
            citecolor=blue,      
            filecolor=magenta,     
            urlcolor=cyan}

\begin{document}

\title{Cost-Aware Dynamic Cloud Workflow Scheduling using Self-Attention and Evolutionary Reinforcement Learning}

\titlerunning{SPN-CWS for CDMWS}  

\author{Ya Shen\inst{}\Envelope \and
Gang Chen\inst{} \and
Hui Ma\inst{} \and
Mengjie Zhang\inst{}
}

\authorrunning{Y. Shen et al.}

\institute{Centre for Data Science and Artificial Intelligence \& School of Engineering and Computer Science, Victoria University of Wellington, New Zealand\\ 
\email{\{ya.shen, aaron.chen, hui.ma, mengjie.zhang\}@ecs.vuw.ac.nz} 
} 

\date{}
\maketitle  

\begin{abstract}
As a key cloud management problem, Cost-aware Dynamic Multi-Workflow Scheduling (CDMWS) aims to assign virtual machine (VM) instances to execute tasks in workflows so as to minimize the total costs, including both the penalties for violating Service Level Agreement (SLA) and the VM rental fees. Powered by deep neural networks, Reinforcement Learning (RL) methods can construct effective scheduling policies for solving CDMWS problems. Traditional policy networks in RL often use basic feedforward architectures to separately determine the suitability of assigning any VM instances, without considering all VMs simultaneously to learn their global information. This paper proposes a novel self-attention policy network for cloud workflow scheduling (SPN-CWS) that captures global information from all VMs. We also develop an Evolution Strategy-based RL (ERL) system to train SPN-CWS reliably and effectively. The trained SPN-CWS can effectively process all candidate VM instances simultaneously to identify the most suitable VM instance to execute every workflow task. Comprehensive experiments show that our method can noticeably outperform several state-of-the-art algorithms on multiple benchmark CDMWS problems.


\end{abstract}


\section{Introduction} \label{section:Introduction}
\begin{figure}[htbp]
\centering
\includegraphics[width=0.55\linewidth]{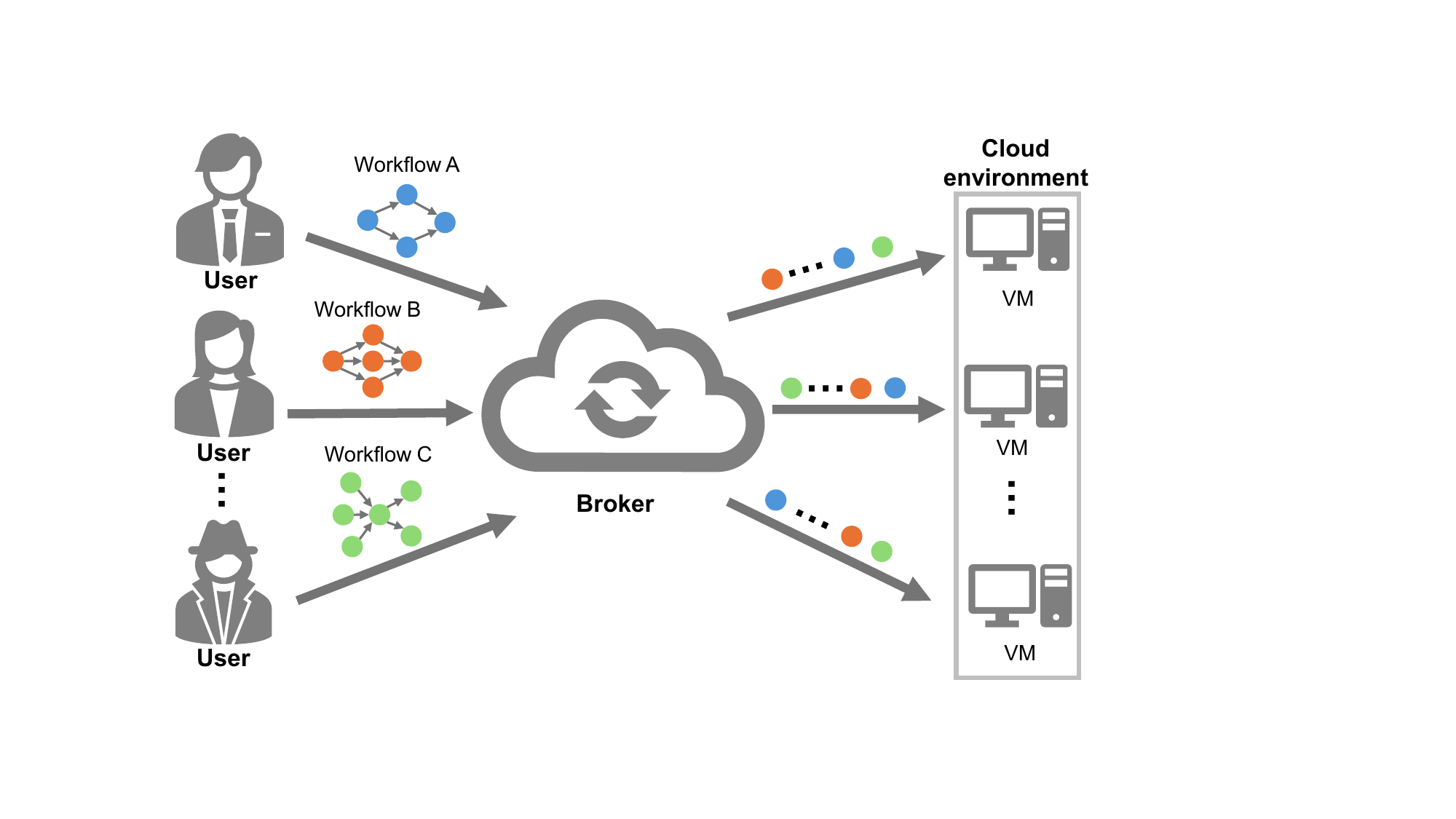} 
\caption{The diagram of the broker performing the workflow scheduling.}
\label{Broker}
\vspace{-5mm}
\end{figure}

Numerous computationally intensive and resource-demanding applications (e.g., weather forecasting, tsunami detection) will be submitted daily as workflows to a \emph{cloud broker} for execution \cite{dong2021workflow,faragardi2019grp,masdari2016towards}. Each workflow comprises a collection of interdependent tasks that must be efficiently processed by using multiple leased virtual machine (VM) instances provided by major cloud providers, such as Amazon EC2 \cite{arabnejad2018budget,huang2022cost}. 
As a primary challenge, the cloud broker must quickly determine the type and number of VMs required to execute these tasks at the lowest possible cost. The broker then leases VM instances to process these workflows. This challenge is widely known as the \emph{workflow scheduling} (WS) problem \cite{huang2022cost}. Fig. \ref{Broker} illustrates how the broker schedules workflows on behalf of its users. Specifically, users submit their workflows to the broker.  
Each workflow is associated with user-defined \emph{Service Level Agreement} (SLA) \cite{wu2013sla,yang2021budget}, requesting the broker to meet the execution deadline to avoid any \emph{SLA violation penalties}. Consequently, the broker must dynamically maintain a desirable trade-off between VM rental fee and \emph{SLA violation penalty}, making the WS problem extremely difficult to solve.

\vspace{-4mm}

This study addresses the \emph{Cost-aware Dynamic Multi-Workflow Scheduling} (CDMWS) problem, where the broker handles multiple dynamically arriving workflows and makes real-time scheduling decisions. Due to the dynamic and combinatorial nature of the CDMWS problem, it is a well-known NP-hard problem \cite{huang2022cost}. To tackle this problem, the broker relies on a scheduling policy to make real-time scheduling decisions and instantly respond to newly arriving workflows. However, the suitability of using any scheduling policy may vary significantly across different problem scenarios \cite{chen2023collaborative,xu2023genetic}. The broker may fail to achieve its performance commitment and cost objectives upon using poorly designed policies \cite{yang2024dual}. Designing scheduling policies is hence a major research issue for dynamic multi-workflow scheduling.

Existing studies on scheduling policies can be classified into three categories: \emph{manual approach}, \emph{genetic programming hyper-heuristic} (GPHH), and \emph{reinforcement learning} (RL) (see Section \ref{section:Related Works}). However, most of those approaches use priority rules to rank individual VMs, without considering the status of other VMs.   
Further, most previously studied policy networks \cite{chen2023collaborative,huang2022cost,jayanetti2024multi,zhou2023deep} adopt basic feedforward architectures with fixed inputs and outputs. They prioritize each candidate VM for task execution based solely on its own features, lacking the ability to consider features across multiple VMs simultaneously, which is essential to accurately capture workflow status and VM relationships and crucial for effective WS.


To tackle this research issue, we have the goal to propose a novel \emph{Self-attention Policy Network for Cloud Workflow Scheduling} (SPN-CWS) to solve the CDMWS problem. In particular, for a given task to be scheduled, information regarding all candidate VMs are fed together into SPN-CWS, which then processes the global information using the self-attention mechanism. Subsequently, SPN-CWS outputs the priority value with respect to each VM based on the global information and task-specific features. The VM with the highest priority is then selected to execute the task. Additionally, we develop an Evolution strategies-based RL \cite{salimans2017evolution} (ERL) system to train SPN-CWS. Compared to gradient-based RL algorithms, ERL is more robust to varied hyperparameter settings and can effectively cope with delayed or sparse rewards \cite{jayanetti2024multi}. Furthermore, ERL can expedite the training process through parallelization. The contributions of our study are summarized below:
\vspace{-0.1cm}
\begin{itemize}
    \item We design a new SPN-CWS policy network  that can effectively utilize global information across all candidate VMs to make informed task scheduling decisions. We adopt the first time in literature the self-attention mechanism in SPN-CWS to scalably handle global relationships among all candidate VMs. 
    \item We develop an ERL system based on our cloud simulator to reliably train SPN-CWS. The trained SPN-CWS is subsequently employed by the broker to execute workflows that arrive dynamically over time.
    \item We conduct comprehensive experiments to thoroughly examine the performance of SPN-CWS and the accompanying ERL system. Experimental results show that the SPN-CWS trained by ERL can notably outperform multiple state-of-the-art approaches for dynamic WS.
\end{itemize}

\vspace{-0.1cm}
The remaining of this paper is organized as follows: Section \ref{section:Related Works} and Section \ref{section:Problem Definition} give the related works and problem definitions of the CDMWS. Sections \ref{section:policy network} and \ref{section:ERL-SPN} describe the proposed SPN-CWS and its training method. Section \ref{section:Experiments Setting} analysis the experiment results and the conclusions of this paper are given in Section \ref{section:Conclusions}. 

\vspace{-0.3cm}
\section{Related Work}\label{section:Related Works}
\vspace{-0.2cm}

Research on dynamic WS is gaining increasing attention, driven by their substantial practical importance in diverse applications \cite{huang2022cost,xu2023genetic,yang2022dual}. Dynamic WS presents a significant challenge, as it requires making real-time scheduling decisions tailored to the current cloud environment where no prior knowledge exists regarding workflows arriving in the future. While previous studies of dynamic WS \cite{chen2018uncertainty,liu2019online} primarily focused on minimizing VM rental costs, a critical aspect often neglected was the trade-off between VM rental expenses and potential SLA violation penalties \cite{faragardi2019grp,wu2017deadline}. In fact, it may prove economically prudent to incur SLA penalties since meeting SLA deadlines often demand for leasing fast but expensive VMs \cite{huang2022cost,xu2023genetic,yang2022dual}. Drawing inspiration from several recent studies \cite{hoseiny2021joint,huang2022cost,yang2022dual}, this paper embarks on an investigation into the CDMWS problem, aiming to learn scheduling policy to jointly optimize the VM rental fees and the SLA violation penalties.

There are three main categories of algorithms for learning scheduling policies for dynamic WS in cloud: manual approach, GPHH and RL. Manual approach \cite{arabnejad2018budget,faragardi2019grp,wu2017deadline} relies on domain experts to design scheduling policies by leveraging their problem-specific knowledge and empirical experiences. However, the designed policies are typically applied to simplified and static problems \cite{huang2022cost,silver2004overview}.

GPHH can automatically design WS policies that are highly adaptive to dynamic WS problems. For example, \cite{yang2022dual} introduced a novel dual-tree policy representation for GPHH. The performance of GPHH is further enhanced with an adaptive mutation operator in \cite{yang2024dual}. A multi-tree GPHH method is also proposed in \cite{xu2023genetic} to solve dynamic WS problems in fog computing environments. However, policies learned by GPHH focus solely on local information, considering each VM's features individually. Without explicitly processing the global information among all VMs (i.e., the operating status of all VMs managed by the broker and the relationships among these VMs), the learned policy may fail to identify suitable VMs for executing some workflow tasks, resulting in increased total cost.

In addition to GPHH, recent studies show promise of using RL methods to design policies for dynamic WS \cite{dong2021workflow,jayanetti2024multi,li2022weighted}. For instance, \cite{jayanetti2024multi} developed an RL system to optimize the utilization of green energy while executing workflow tasks. An approach based on deep Q-networks was introduced by \cite{wang2019multi} with the objective of optimizing both workflow makespan and user costs. In \cite{chen2023collaborative}, a RL-based collaborative scheduling method is proposed to achieve heterogeneous WS in cloud, aiming to improve overall service quality. 
However, most of the existing policy networks trained by RL adopt simple feedforward architectures with fixed inputs and outputs. Similar to GPHH, they were designed to process features related to each VM separately, lacking the capability of processing global information across all candidate VMs. 

To address this issue, we design the first time in literature a new policy network (SPN-CWS) to simultaneously process global information among an arbitrary number of candidate VMs. Meanwhile, to ensure that the training of SPN-CWS is robust to hyperparameter settings and reward functions, we develop an ERL system to train SPN-CWS using simulated WS problems.

\vspace{-0.2cm}
\section{Problem Definition} \label{section:Problem Definition}
\begin{figure}[htbp]
\centering
\includegraphics[width=0.7\linewidth]{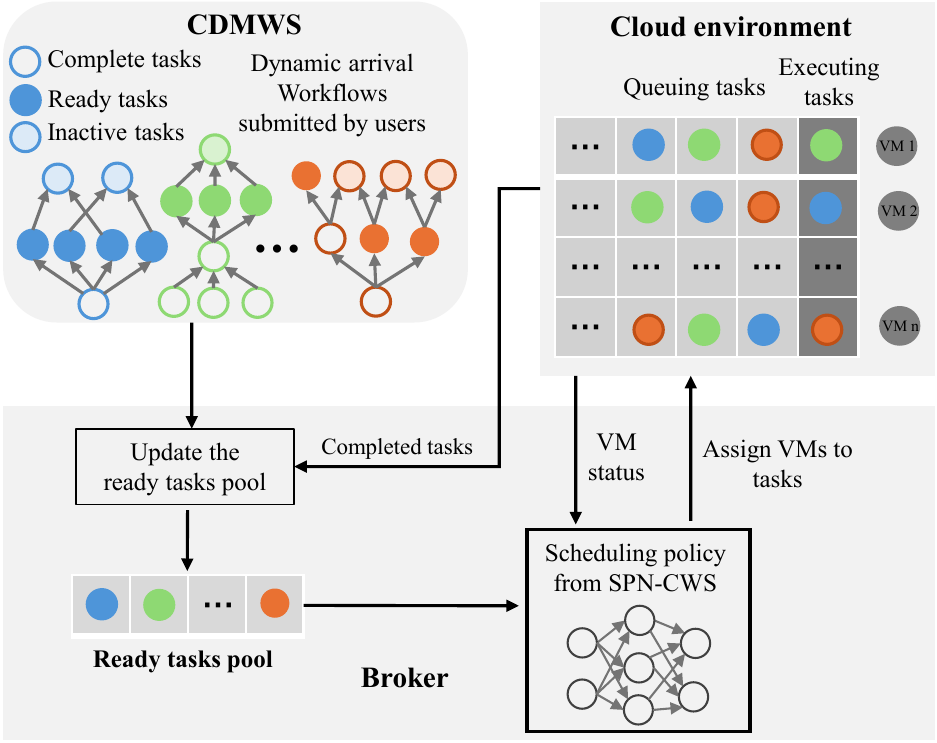} 
\caption{The diagram of the scheduling of CDMWS.}
\label{CDMWS}
\end{figure}

\vspace{-0.1cm}
In this section, we formally describe the Cost-aware Dynamic Multi-Workflow Scheduling problem (CDMWS) as illustrated in Fig. \ref{CDMWS}. This problem is centered around a \emph{broker} that is responsible for dynamically scheduling workflow execution using leased VMs in cloud to minimize the total cost, including both the VM Rental fees and the SLA violation penalties.

\vspace{0.2cm}
\textbf{Workflow model:} In CDMWS, we assume that a sequence of workflows are dynamically submitted by users during $T$: $W(T)=\{w_{i}|i=1,2,... \}$. Each workflow $w_i$ contains a set of tasks connected by a \emph{Directed Acyclic Graph} (DAG). A workflow $w$ can be specifically formulated as follows:
\begin{equation}
\label{eq:Workflow Model}
    w = \left[DAG(w),AT(w),DL(w),\beta(w)|{w \in W(T)} \right]
\end{equation}
where $\beta(w)$ is the user specified SLA penalty rate \cite{youn2017cloud} (see Eq. \eqref{eq:SLA penalty}). $AT(w)$ is the arrival time of $w$, and $DL(w)$ is the user specified SLA deadline (see Eq. \eqref{eq:SLA deadline}). $DAG(w)$ provides a directed acyclic graph of tasks, representing all the tasks to be executed as part of $w$ as nodes and showing how these tasks are connected together through the directed edges in $DAG(w)$. $DAG(w)$ can be formulated as below:
\begin{equation}
\label{eq:DAG}
    DAG(w) = \left[Task(w),Edge(w)\right]
\end{equation}
where $Task(w)$ gives the set of tasks in workflow $w$. Each task $t\in Task(w)$ has its size determined as $Size(t)$ that quantifies the total amount of computation required for executing it. Each direct edge $(t,t')\in Edge(w)$ connects one task $t\in Task(w)$ to another task $t'\in Task(w)$, indicating that $t$ is a \emph{predecessor} task of $t'$ and $t'$ is a \emph{successor} task of $t$. Specifically, any task $t\in Task(w)$ will become \emph{ready} for execution only when all of its predecessor tasks are completed or no predecessor tasks exist. 

\vspace{0.2cm}
\textbf{Cloud environment:} the cloud environment contains a set of VMs with varied VM types. In this study, the types of VM are limited, but the quantity available for each type is unlimited, meaning that the broker can lease an arbitrary number of VM instances of any type on demand. A VM instance denoted as $v$ in the cloud can be characterized as follows: 
\begin{equation}
\label{eq:The cloud environment}
    v = \left[Type(v),Capacity(v),Price(v)\right]
\end{equation}
where $Type(v)$ indicates the VM type; $Capacity(v)$ gives the computation capability per time unit of $v$ \cite{faragardi2019grp} ; $Price(v)$ is the rental fee per hour incurred as a result of using/leasing $v$. According to several existing works \cite{faragardi2019grp,huang2022cost,yang2022dual}, the rental time for a fraction of an hour is charged as one hour.

\vspace{0.2cm}
\textbf{Execution time:} Let the VM instance chosen by policy $\pi$ to execute task $t\in Task(w)$ of workflow $w$ be denoted as $v_{t,w,\pi}$. The required execution time of $t$, i.e., $EXT(t,w,\pi)$, is determined as:
\vspace{-2mm}
\begin{equation}
\label{eq:Execution time}
    EXT(t,w,\pi) = \frac{Size(t)}{Capacity(v_{t,w,\pi})}
\end{equation}
We further denote $AVM(v|t,w,\pi)$ as the set of all candidate VM instances that can be used to execute task $t\in Task(w)$ of workflow $w$, including all the currently rented VMs as well as VMs that can be newly leased from the cloud. Clearly $v_{t,w,\pi}\in AVM(v|t,w,\pi)$.

In this study, workflows arrive at the broker dynamically across time, demanding $\pi$ to make its scheduling decisions in real time. In order to reduce the VM rental fees and the SLA penalty, every ready task must be scheduled for execution immediately whenever it becomes ready~\cite{huang2022cost}. The following formulation captures the completion time of a $ready$ task $t\in Task(w)$ of workflow $w$:
\vspace{-2mm}
\begin{equation}
\label{eq:Completion time task}
    CT(t,w,\pi) = ST(t,w,\pi)+EXT(t,w,\pi)
\end{equation}
where $CT(t,w,\pi)$ and $ST(t,w,\pi)$ are the completion time and start time of $t\in Task(w)$ upon using $\pi$, respectively. Hence, the completion time of workflow $w$ under policy $\pi$ is:
\vspace{-2mm}
\begin{equation}
\label{eq:Completion time workflow}
    WCT(w,\pi) = \max_{t\in Task(w)} CT(t,w,\pi)
\end{equation}

\vspace{0.2cm}
\textbf{VM rental fees:} CDMWS focuses on processing multiple dynamically arriving workflows, each consisting of a collection of tasks to be executed on multiple VMs. When all workflows are completed, the total rental fees of all leased VMs will be calculated. We use $T$ to represent the full operation period associated with a CDMWS problem. This time period starts from time $t_{s}$ and ends at time $t_{e}$. The duration of $T$ depends on the scheduling policy $\pi$ used to execute all workflow tasks. During $T$, each VM instance $v$ is used for a certain time period, as defined below:
\vspace{-2mm}
\begin{equation}
\label{eq:VM Rental period}
    RP(v,\pi,T) = \left[ t_{s}(v,\pi,T),t_{e}(v,\pi,T)\right]
\end{equation}
where $t_{s}(v,\pi,T)\geq t_s$ is the start time for using $v$, and $t_{e}(v,k,\pi,T)\leq t_e$ is the time at which $v$ is no longer used. Thus, the total rental fees of all VMs under the scheduling policy $\pi$ during $T$ can be calculated as follows:
\vspace{-2mm}
\begin{equation}
\label{eq: Total Rental Fees}
    VMFee(\pi,T) = \sum_{v \in Set(\pi,T)} \left( Price(v) \times \left\lceil \frac{t_{e}(v,\pi,T)-t_{s}(v,\pi,T)}{3600} \right\rceil \right)
\end{equation}
where $Set(\pi,T)$ refers to the set of all VM instances leased to execute workflow tasks during $T$. The ceiling function in Eq. \eqref{eq: Total Rental Fees} converts the time period that a VM instance $v$ is used to its corresponding renting time in integer hours.

\vspace{0.2cm}
\textbf{SLA penalty:} Apart from VM rental fees, the SLA penalty presents a major source of cost to be minimized in CDMWS. According to \cite{huang2022cost,wu2013sla,yang2022dual}, the SLA penalty of a workflow $w$ can be calculated by the following:
\vspace{-2mm}
\begin{equation}
\label{eq:SLA penalty}
    SLA_{Penalty}(w,\pi) = \beta(w) \times \max \{0,[WCT(w,\pi)-DL(w)]  \}
\end{equation}
In Eq. \eqref{eq:SLA penalty}, the SLA deadline specified by users, i.e., $DL(w)$, is determined as:
\vspace{-2mm}
\begin{equation}
\label{eq:SLA deadline}
    DL(w) = AT(w)+\gamma \times MinMakespan(w)
\end{equation}
where $MinMakespan(w)$ is the theoretical shortest time duration required for processing $w$, achievable by using the fastest VM to process all tasks of workflow $w$ without any delay. $\gamma$ is the relaxation coefficient. Increasing $\gamma$ results in more relaxed deadline for $w$.


In CDMWS, we aim to find an effective (optimal) scheduling policy $\pi$ to minimize the total cost, as formulated below:
\vspace{-2mm}
\begin{equation}
\label{eq:Objective of CDMWS}
    \arg\min_{\pi} TotalCost(\pi)= \arg\min_{\pi} \left\{ VMFee(\pi,T)+\sum_{w \in W(T)} SLA_{Penalty}(w,\pi) \right\}
\end{equation}

\section{Self-Attention Policy Network for Cloud Workflow Scheduling (SPN-CWS)} \label{section:policy network}
\subsection{State information} \label{subsection:State information}

The scheduling policy plays a critical role for CDMWS. Whenever a task (i.e., $rt$) of any workflow (i.e., $w_{rt}$) becomes ready for execution at time $t$, $\pi$ is utilized to process the state information obtained from the CDMWS problem as its input, and then produces a VM selected to execute $rt$ as its action output. Therefore, we first introduce the state representation in association with the CDMWS problem being solved, which will serve as the input of SPN-CWS. 

In line with \cite{huang2022cost,yang2022dual}, we design the state representation by considering both the \emph{task-info} of the ready task $rt$ and the \emph{VM-info} of all the VMs in $AVM(v|rt,w_{rt},\pi)$ that can be utilized for executing $rt$. We specifically list all the features employed to build the state input with respect to \emph{task-info} and \emph{VM-info} in Table \ref{Table:task_vm_info}. Whenever there is a task $rt$ ready to be processed, the \emph{task-info} of $rt$ and the \emph{VM-info} of all VM instances in $AVM(v|rt,w_{rt},\pi)$ jointly form the representation of the current state as the input of SPN-CWS.

\begin{table}[htbp]
\vspace{-8mm}
\centering
\caption{Task information (\emph{task-info}) and VM information (\emph{VM-info}).}
\vspace{-2mm}
\label{Table:task_vm_info}
\begin{center}
\scalebox{0.9}{
\begin{tabular}{|>{\centering\arraybackslash}m{1.25cm}|>{\raggedright\arraybackslash}m{2cm}|>{\raggedright\arraybackslash}m{10cm}|}
\hline
\multirow{3}{*}{\centering\emph{task-info}} & \textbf{\emph{Number of Successors}} & It gives the number of tasks that depend on $rt$ in $w_{rt}$, i.e., $|Suc(w_{rt},rt)|$, where $|\cdot|$ stands for the set cardinality. \\ \cline{2-3}
& \textbf{\emph{Completion ratio}} & It calculates the workflow completion ratio of $w_{rt}$ through $Com_{tasks}/Total_{tasks}$, where $Com_{tasks}$ is the number of completed tasks in $w_{rt}$ and $Total_{tasks}$ is the total number of tasks in $w_{rt}$. \\ \cline{2-3}
& \textbf{\emph{Arrival rate}} & It estimates the workflow arrival rate for future workflows according to the current execution situation of $w_{rt}$ \cite{huang2022cost}. \\ \hline
\multirow{4}{*}{\centering\emph{VM-info}} & \textbf{\emph{Task deadline}} & It indicates whether using the VM  can meet the task deadline of $rt$. The task deadline is calculated using the method in \cite{wu2017deadline}. Depending on the task size, the workflow deadline is assigned to each task. Large size task will be assigned a larger task deadline. \\ \cline{2-3}
& \textbf{\emph{Incurred cost}} & It calculates the corresponding VM rental fee and \textbf{\emph{Task deadline}} violation penalty to be incurred upon using the VM to execute $rt$. \\ \cline{2-3}
& \textbf{\emph{Remaining time}} & It counts the remaining VM rental time after executing $rt$ on this VM. \\ \cline{2-3}
& \textbf{\emph{Fittest VM}} & It indicates whether the VM being considered for executing $rt$ has the lowest \textbf{\emph{Incurred cost}} among all candidate VMs while meeting the \textbf{\emph{Task deadline}}. \\ \hline
\end{tabular}}
\end{center}
\vspace{-5mm}
\end{table}

\subsection{Architecture Design of SPN-CWS}

\begin{figure}[htbp]
\centering
\includegraphics[width=\linewidth]{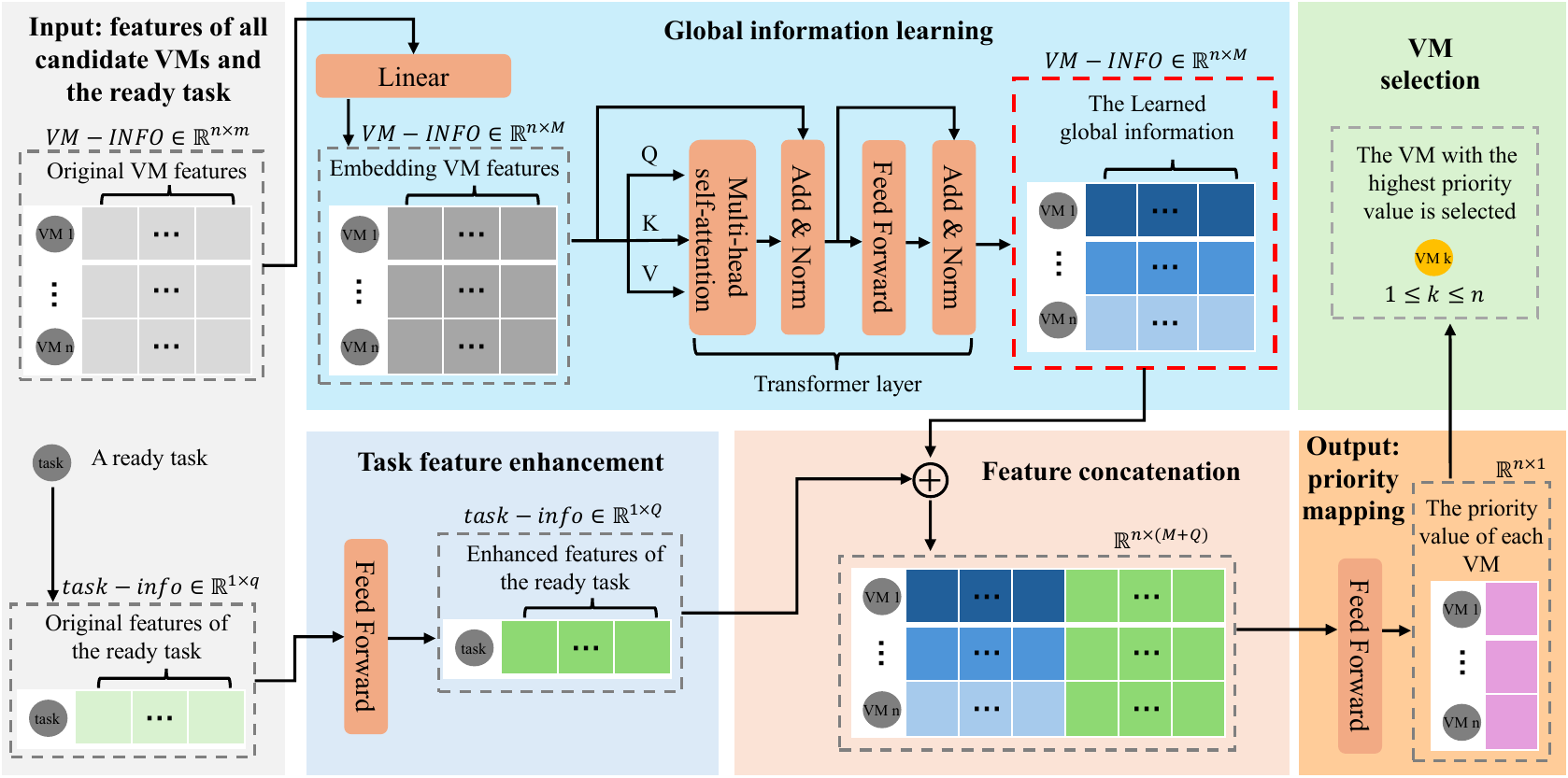} 
\caption{The structure of SPN-CWS.}
\label{policy network}
\end{figure}

To process global information across all VMs in $AVM(v|rt,w_{rt},\pi)$ for effective scheduling of the ready task $rt$, we design a novel SPN-CWS deep model for the scheduling policy $\pi$ in CDMWS, as shown in Fig. \ref{policy network}. Specifically, we introduce the time-tested \emph{Multi-Head Self-Attention} mechanism (MHSA) \cite{vaswani2017attention}, as highlighted in the \textbf{\emph{Global information learning}} component of Fig. \ref{policy network}, to enhance the capability for policy $\pi$ to process global VM information effectively and scalably (see the detailed experiments in Section \ref{subsection:Simulation results}). SPN-CWS mainly consists of five key components as described below.
\vspace{-3mm}
\vspace{-4mm}

\textbf{\emph{Global information learning}}: This component aims to learn global information among VMs. Firstly, we combine the \emph{VM-info} of all VMs in $AVM(v|rt,w_{rt},\pi)$ to construct the \emph{VM-INFO} indicated in Fig. \ref{policy network}. Then, we employ a linear layer to map \emph{VM-INFO} from the original $\mathbb{R}^{n \times m}$ space to a high-dimensional $\mathbb{R}^{n\times M}$ space ($M>m$ and $n=|AVM(v|rt,w_{rt},\pi)|$, i.e., the total number of VM instances in $AVM(v|rt,w_{rt},\pi)$). Then, the Transformer layer processes the projected VM features to learn the relationships among all VMs. In more details, first we use MHSA to compute a weighted sum of all VM features, where the weights are determined by the similarity between every pair of VMs. The weighted features are further processed by Add \& Norm and Feed-Forward Networks (FFN) to learn more complex feature representations. MHSA allows features of each VM instance to attend to features of other VM instances, effectively capturing inter-VM relationships and allowing SPN-CWS to focus on the most relevant VM instances.

\textbf{\emph{Task feature enhancement}}: This component processes and transforms task features. In this component, a FFN is used to map \emph{task-info} from the $\mathbb{R}^{1 \times q}$ space to a high-dimensional $\mathbb{R}^{1 \times Q}$ space ($Q>q$) to build high-level feature representations of the ready task $rt$ being scheduled. FFN refines the task features, ensuring that the task representation captures important information that is essential for the subsequent concatenation with VM features.

\textbf{\emph{Feature concatenation}}: This component concatenates task features of the ready task with the features from each VM instance. In order to perform priority mapping on every VM instance, in this component, the processed \emph{task-info} is concatenated separately with the \emph{VM-INFO} associated with each candidate VM instance.

\textbf{\emph{Priority mapping}}: This component maps concatenated task and VM features to priority values that dictate the importance of using any specific VM instance to execute the ready task. For example, using the \emph{VM-INFO} features with respect to $v_1$ and the features in the \emph{task-info}, the priority value of VM instance $v_1$ is determined through a FFN included in this component, as shown in Fig. \ref{policy network}.

\textbf{\emph{VM Selection}}: This component selects the VM instance with the highest priority value to execute the ready task $rt$.

\vspace{-1mm}
\section{Training SPN-CWS via ERL} \label{section:ERL-SPN}
The performance of gradient-based RL can be highly sensitive to its hyperparameter settings as well as the design of the reward function. ERL can alleviate these issues and achieve robust learning performance \cite{ajani2022adaptive,khadka2018evolution,salimans2017evolution}. Therefore, in this study, we develop an ERL system based on our cloud simulator to train SPN-CWS designed above for CDMWS, which is denoted as $\pi_{sc}$. The pseudo-code of the training algorithm is presented in Algorithm \ref{algorithm:ERL-SPN}. Specifically, the training process of $\pi_{sc}$ using ERL is described as follows:
\begin{enumerate}
\item Let $\hat{\theta}$ denote the current policy parameter of $\pi_{sc}$. During each generation (i.e., each iteration of the ERL), ERL samples a population of $N$ individuals, where each individual ($\theta_{i}|i=1,2,...N$) is sampled from an isotropic multivariate Gaussian distribution with $\hat{\theta}$ as its mean vector and $\sigma^{2}I$ as its co-variance matrix. Hence, $\theta_{i} \sim \mathcal{N}(\hat{\theta}, \sigma^{2}I)$, which is equivalent to $\theta_{i}=\hat{\theta}+\sigma\epsilon_{i}$ with $\epsilon_{i} \sim \mathcal{N}(0, I)$. $\theta_{i}$ indicates the parameters of $\pi_{i}$.

\item The fitness value of $\theta_{i}$ (i.e., $F(\theta_{i})$) equals the total cost (as described in Section \ref{section:Problem Definition}) achieved by using $\pi_{i}$ to solve any CDMWS problem used for training. Since ERL aims to learn the policy parameters that minimize the total cost, we define the fitness function as follows:
\begin{equation}
\label{eq:Fitness function}
    F(\theta_{i}) = F(\hat{\theta}+\sigma\epsilon_{i}) = -TotalCost(\pi_{i})
\end{equation}

\item With Eq. \eqref{eq:Fitness function} as the objective function, ERL updates $\hat{\theta}$ to maximize the expected value of the objective function (i.e., $\mathbb{E}_{\theta_{i} \sim \mathcal{N}(\hat{\theta}, \sigma^{2}I)} F(\theta_{i})$), thereby minimizing the total costs. Specifically, ERL updates $\hat{\theta}$ using policy gradients estimated below \cite{salimans2017evolution}:

\begin{align}
\label{eq:gradient estimator}
\nabla_{\hat{\theta}} \mathbb{E}_{\theta_{i} \sim \mathcal{N}(\hat{\theta}, \sigma^{2}I)} F(\theta_{i}) &= \nabla_{\hat{\theta}} \mathbb{E}_{\epsilon_{i} \sim \mathcal{N}(0, I)} F(\hat{\theta}+\sigma\epsilon_{i}) \\
& = \frac{1}{\sigma} \mathbb{E}_{\epsilon_{i} \sim \mathcal{N}(0, I)}\{F(\hat{\theta}+\sigma\epsilon_{i})\epsilon_{i}\} \nonumber 
\end{align}
\end{enumerate}

\vspace{-5mm}
\begin{algorithm}[htbp]
\SetAlgoLined
\SetKwInOut{Input}{\textbf{Input}}
\SetKwInOut{Output}{\textbf{Output}}
\Input{Population size: $N$, max number of generation: $Gen$, initial parameters of $\pi_{sc}$: $\hat{\theta}$, initial learning rate: $\alpha$, and the Gaussian standard noise deviation: $\sigma$}
\Output{Scheduling policy: $\pi_{sc}$ (the trained SPN-CWS)}
\BlankLine
\textbf{While} the current number of generation $<=$ $Gen$: \textbf{do} \\
\hspace{0.5cm} Randomly generate a CDMWS training problem from our simulator: $Pro$ \\
\hspace{0.5cm} \textbf{For} each individual ($i$=1,2,...) \textbf{in} $N$: \textbf{do} \\
\hspace{1cm} Sample a $\epsilon_{i} \sim \mathcal{N}(0, I)$ \\
\hspace{1cm} The parameters of $\pi_{i}$ represented by individual $i$: $\theta_{i}=\hat{\theta}+\sigma\epsilon_{i}$ \\
\hspace{1cm} Evaluate the fitness value of $F(\theta_{i})$ using Eq. \eqref{eq:Fitness function} based on $Pro$  \\
\hspace{0.5cm} \textbf{End for} \\
\hspace{0.5cm} Estimate the policy gradient $\nabla_{\hat{\theta}} \mathbb{E}_{\theta_{i} \sim \mathcal{N}(\hat{\theta}, \sigma^{2}I)} F(\theta_{i})$ using Eq. \eqref{eq:gradient estimator}  \\
\hspace{0.5cm} Update parameters of $\pi_{sc}$: $\hat{\theta} \leftarrow$ $\hat{\theta} + \alpha \nabla_{\hat{\theta}} \mathbb{E}_{\theta_{i} \sim \mathcal{N}(\hat{\theta}, \sigma^{2}I)} F(\theta_{i})$ \\
\textbf{End while}
\caption{ERL for training SPN-CWS ($\pi_{sc}$)}
\label{algorithm:ERL-SPN}
\end{algorithm}

\section{Experiments} \label{section:Experiments Setting}
\subsection{Simulation Environment Configuration} \label{subsection:Simulation Environment Configuration}
In this section, we present the CDMWS simulation environment for training SPN-CWS using ERL as well as all the competing algorithms. The simulation environment is specifically described as follows.

\textbf{VM types and workflow patterns:} We configure the VMs used in CDMWS based on Amazon EC2\footnote{https://aws.amazon.com/ec2/pricing/on-demand/}. In line with existing studies \cite{huang2022cost,yang2024dual,yang2022dual}, we use six types of VMs with their respective configurations summarized in Table \ref{Table:1}. Following \cite{huang2022cost,yang2024dual}, each type of VM can be leased on demand with unlimited number of instances. As summarized in Table \ref{Table:2}, the workflows experimented consist of four popular patterns (i.e., CyberShake, Montage, Inspiral and SIPHT) that are commonly used in recent studies \cite{huang2022cost,xu2023genetic,yang2024dual}. Workflows of the same pattern become more complex and difficult to solve as the number of tasks in the workflows increases. Based on the workflow size (i.e., number of workflow tasks), workflows are grouped into three categories: \emph{Small}, \emph{Medium}, and \emph{Large}.
\begin{table}[htbp]
\vspace{-8mm}
\centering
\caption{All VM types used in this study.}
\vspace{-2mm}
\label{Table:1}
\begin{center}
\scalebox{0.8}{
\begin{tabular}{|c|c|c|c|c|c|}
\hline
\textbf{VM name} & \textbf{vCPU} & \textbf{Memory (GB)} & \textbf{Cost (\$ per hour)} \\ \hline
m5.large     & 2    & 8     & 0.096 \\ \hline
m5.xlarge    & 4    & 16    & 0.192 \\ \hline
m5.2xlarge   & 8    & 32    & 0.384 \\ \hline
m5.4xlarge   & 16   & 64    & 0.768 \\ \hline
m5.8xlarge   & 32   & 128   & 1.536 \\ \hline
m5.12xlarge  & 48   & 192   & 2.304 \\ \hline
\end{tabular}}
\end{center}
\vspace{-10mm}
\end{table}

\textbf{CDMWS problem instance:} 
Following \cite{huang2022cost,yang2024dual}, each CDMWS problem instance consists of 30 randomly sampled workflows corresponding to all the four workflow patterns in Table \ref{Table:2}. All workflows in this study are dynamically generated according to a Poisson distribution with $\lambda = 0.01$ to simulate workflows submitted by users across time. We set the SLA deadline coefficient\footnote{ES-RL and SPN-CWS are trained with $\gamma = 5$ and tested with $\gamma \in \{1.00, 1.25, 1.50,$ $1.75, 2.00, 2.25\}$ to evaluate their performance under tight SLA deadline coefficients.} in Eq. \eqref{eq:SLA deadline} as: $\gamma \in \{1.00, 1.25, 1.50, 1.75, 2.00, 2.25\}$. SLA deadlines become more relaxed as $\gamma$ increases, enabling the broker to rent cheaper VMs to execute newly arrived workflows. $\beta(w)$ in Eqs. \eqref{eq:Workflow Model} and \eqref{eq:SLA penalty} is set to $\$0.24/hour$ according to \cite{huang2022cost,yang2024dual}.
\begin{table}[htbp]
\vspace{-8mm}
\centering
\caption{The three workflow sets used in this study.}
\label{Table:2}
\begin{center}
\scalebox{0.8}{
\begin{tabular}{|c|c|c|c|c|}
\hline
\textbf{Workflow set} & \multicolumn{4}{c|}{\textbf{Name of pattern(number of task)}} \\ \hline
\emph{Small}    & CyberShake(30)   & Montage(25)   & Inspiral(30)   & SIPHT(30)  \\ \hline
\emph{Medium}   & CyberShake(50)   & Montage(50)   & Inspiral(50)   & SIPHT(60)  \\ \hline
\emph{Large}    & CyberShake(100)  & Montage(100)  & Inspiral(100)  & SIPHT(100)  \\ \hline
\end{tabular}}
\end{center}
\vspace{-10mm}
\end{table}

\textbf{CDMWS problem scenarios:} Small-scenario CDMWS problem instances are generated from \emph{small} workflow set. Meanwhile, medium- and large-scenario CDMWS problem instances are generated from \emph{Medium} and \emph{Large} workflow sets, respectively. For each generation of Algorithm \ref{algorithm:ERL-SPN}, we sample a small-scenario CDMWS problem instance for fitness evaluation. During testing, we use 30 small-scenario, 30 medium-scenario, and 30 large-scenario CDMWS instances to jointly evaluate the performance of the trained SPN-CWS. Notably, the medium- and large-scenario CDMWS problems instances are not used during training for computation efficiency reasons. They are only employed to assess the generalization capability of SPN-CWS during testing. In each test scenario, the scheduling policy's performance is calculated based on the average of 30 CDMWS problem instances.

\textbf{Baseline algorithms:} Four competing methods are included in our experiments. Among them, \textbf{ProLis} \cite{wu2017deadline} and \textbf{GRP-HEFT} \cite{faragardi2019grp} are two popular heuristic methods designed manually by domain experts and are frequently studied in existing works \cite{dong2021workflow,xu2023genetic,yang2024dual}. \textbf{DSGP} \cite{escott2020genetic} is a GPHH-based approach that learns a heuristic rule as scheduling policy through evolutionary search in the hyper-heuristic space. \textbf{ES-RL} \cite{huang2022cost} adopts an RL-based algorithm designed by OpenAI\footnote{https://github.com/openai} to train a scheduling policy modeled as a feedforward neural network.

\textbf{Parameter settings:} The parameter settings of all competing algorithms strictly follow their original papers. For Algorithm \ref{algorithm:ERL-SPN}, $N$, $Gen$, $\alpha$, and $\sigma$ are set to 40, 3000, 0.01, and 0.05, respectively, which are identical to the settings adopted in \cite{huang2022cost} for a fair comparison. According to Table \ref{Table:task_vm_info}, $m$ and $q$ of Fig. \ref{policy network} are 4 and 3, respectively, while $M$ and $Q$ are set to 16 in this study. In SPN-CWS, the size of feedforward hidden layers in the \emph{Global information learning}, \emph{Task feature enhancement}, and \emph{Priority mapping} components are set to 64, 32, and 32, respectively, with \emph{ReLU} as the activation function. 

\subsection{Main Results} \label{subsection:Simulation results}
The performance of all competing algorithms are summarized in Table \ref{Table:Main results}, which clearly indicates that SPN-CWS significantly outperforms ProLis and GRP-HEFT across all scenarios, confirming the importance of designing scheduling policies automatically. Furthermore, compared to DSGP and ES-RL, thanks to its capability of processing global information among all candidate VMs, SPN-CWS achieved significantly better overall performance. Particularly, on majority of medium and large scenarios, SPN-CWS significantly outperforms DSGP and ES-RL, with scenario \(\langle2.00, M\rangle\) as the only exception. This shows that SPN-CWS, while being trained on small problems, can achieve reliable generalization performance on large problems.

Meanwhile, we have bolded the best performance results in Table \ref{Table:Main results}, which have been thoroughly verified through the Wilcoxon ranked sum test. The corresponding $p$-values associated with the bolded results are consistently less than 0.05. For example, on the scenario \(\langle1.50, M\rangle\), SPN-CWS can manage to reduce the total costs by 5.29\% and 56.9\% respectively, compared to the total costs realized by DSGP and ES-RL. 

Moreover, Table \ref{Table:Main results} reveals that, upon increasing $\gamma$, the total costs achieved by SPN-CWS consistently exhibit a downward trend across all test scenarios. This suggests that a more relaxed SLA deadline coefficient encourages SPN-CWS to utilize cheaper VM instances, thereby effectively lowering the total costs.
\begin{table}[htbp]
\vspace{-8mm}
\centering
\caption{Average (standard deviation) total cost of each algorithm over 30 independent runs.}
\vspace{-2mm}
\label{Table:Main results}
\begin{center}
\begin{threeparttable}
\scalebox{0.75}{
\begin{tabular}{|c|c|c|c|c|c|}
\hline
Scenarios & ProLis \cite{wu2017deadline} & GRP-HEFT \cite{faragardi2019grp} & DSGP \cite{escott2020genetic} & ES-RL \cite{huang2022cost} & SPN-CWS (our) \\ \hline
\(\langle 1.00, S \rangle\) & 773.69   & 1685.01\ssmall(-)  & \textbf{142.88}(16.35)\ssmall(+)(+)  & 215.83(50.34)\ssmall(+)(+)(-)    & \textbf{145.53}(14.96)\ssmall(+)(+)($\approx$)(+)  \\ 
\(\langle 1.00, M \rangle\) & 1829.25  & 2867.64\ssmall(-)  & 244.14(66.39)\ssmall(+)(+)           & 456.87(125.70)\ssmall(+)(+)(-)   & \textbf{238.68}(51.38)\ssmall(+)(+)(+)(+)          \\ 
\(\langle 1.00, L \rangle\) & 3641.79  & 5873.20\ssmall(-)  & 422.49(96.27)\ssmall(+)(+)           & 1199.20(264.65)\ssmall(+)(+)(-)  & \textbf{348.19}(87.92)\ssmall(+)(+)(+)(+)          \\ \hline
\(\langle 1.25, S \rangle\) & 923.15   & 1967.31\ssmall(-)  & \textbf{128.24}(11.12)\ssmall(+)(+)  & 301.46(77.59)\ssmall(+)(+)(-)    & 131.05(12.67)\ssmall(+)(+)(-)(+)                    \\ 
\(\langle 1.25, M \rangle\) & 2068.78  & 3578.04\ssmall(-)  & 232.34(57.15)\ssmall(+)(+)           & 568.56(173.47)\ssmall(+)(+)(-)   & \textbf{212.72}(35.50)\ssmall(+)(+)(+)(+)           \\ 
\(\langle 1.25, L \rangle\) & 4213.29  & 6868.22\ssmall(-)  & 438.23(80.27)\ssmall(+)(+)           & 1278.11(287.33)\ssmall(+)(+)(-)  & \textbf{348.33}(83.90)\ssmall(+)(+)(+)(+)           \\ \hline
\(\langle 1.50, S \rangle\) & 901.40   & 1963.39\ssmall(-)  & \textbf{126.12}(14.49)\ssmall(+)(+)  & 249.40(54.74)\ssmall(+)(+)(-)    & \textbf{127.76}(12.18)\ssmall(+)(+)($\approx$)(+)           \\ 
\(\langle 1.50, M \rangle\) & 2035.52  & 3567.97\ssmall(-)  & 224.21(47.68)\ssmall(+)(+)           & 492.13(164.01)\ssmall(+)(+)(-)   & \textbf{212.34}(34.98)\ssmall(+)(+)(+)(+)          \\ 
\(\langle 1.50, L \rangle\) & 4066.84  & 6863.62\ssmall(-)  & 405.98(94.18)\ssmall(+)(+)           & 913.00(265.89)\ssmall(+)(+)(-)   & \textbf{343.76}(83.37) \ssmall(+)(+)(+)(+)          \\ \hline
\(\langle 1.75, S \rangle\) & 804.33   & 1959.71\ssmall(-)  & \textbf{123.14}(13.49)\ssmall(+)(+)  & 227.23(46.17)\ssmall(+)(+)(-)    & \textbf{122.58}(10.14)\ssmall(+)(+)($\approx$)\ssmall(+)  \\ 
\(\langle 1.75, M \rangle\) & 1977.10  & 3560.22\ssmall(-)  & 210.46(37.68)\ssmall(+)(+)           & 455.73(136.39)\ssmall(+)(+)(-)   & \textbf{209.54}(33.87)\ssmall(+)(+)(+)(+)          \\ 
\(\langle 1.75, L \rangle\) & 3898.42  & 6854.40\ssmall(-)  & 398.29(84.18)\ssmall(+)(+)           & 818.91(259.21)\ssmall(+)(+)(-)   & \textbf{348.22}(80.40)\ssmall(+)(+)(+)(+)          \\ \hline
\(\langle 2.00, S \rangle\) & 789.71   & 1957.25\ssmall(-)  & \textbf{120.74}(12.47)\ssmall(+)(+)  & 206.94(39.40)\ssmall(+)(+)(-)    & \textbf{119.17}(9.73)\ssmall(+)(+)($\approx$)(+)   \\ 
\(\langle 2.00, M \rangle\) & 1921.42  & 3554.07\ssmall(-)  & \textbf{204.36}(37.68)\ssmall(+)(+)  & 409.84(135.83)\ssmall(+)(+)(-)   & \textbf{205.62}(32.80)\ssmall(+)(+)($\approx$)\ssmall(+)             \\ 
\(\langle 2.00, L \rangle\) & 4024.03  & 6847.49\ssmall(-)  & 388.64(104.18)\ssmall(+)(+)          & 748.58(246.18)\ssmall(+)(+)(-)   & \textbf{349.51}(95.20)\ssmall(+)(+)(+)(+)          \\ \hline
\(\langle 2.25, S \rangle\) & 710.18   & 1954.71\ssmall(-)  & \textbf{115.28}(13.13)\ssmall(+)(+)  & 184.78(31.46)\ssmall(+)(+)(-)    & \textbf{116.93}(10.61)\ssmall(+)(+)($\approx$)(+)          \\ 
\(\langle 2.25, M \rangle\) & 1924.27  & 3551.31\ssmall(-)  & 197.96(35.28)\ssmall(+)(+)           & 362.83(144.59)\ssmall(+)(+)(-)   & \textbf{194.06}(27.23)\ssmall(+)(+)(+)(+)          \\ 
\(\langle 2.25, L \rangle\) & 3957.08  & 6842.88\ssmall(-)  & 373.35(94.76)\ssmall(+)(+)           & 611.03(218.16)\ssmall(+)(+)(-)   & \textbf{343.97}(78.32)\ssmall(+)(+)(+)(+)          \\ \hline
\end{tabular}}
\begin{tablenotes}
\tiny
\begin{minipage}{\linewidth}
\item[*] \(\langle 1.00, S \rangle\) denotes that \(\gamma = 1.00\) and algorithms are tested on the small-scenario CDMWS instance.
\item[*]  ProLis and GRP-HEFT are deterministic heuristics, therefore have no standard deviation.
\item[*] (+), (-) or ($\approx$) indicates that the result is significantly better, worse or equivalent to the corresponding \\algorithm based on Wilcoxon test with a significance level of 0.05.
\end{minipage}
\end{tablenotes}
\end{threeparttable}
\end{center}
\vspace{-8mm}
\end{table}

\begin{figure}[htbp]
\vspace{-3mm}
\centering
\includegraphics[width=0.9\linewidth]{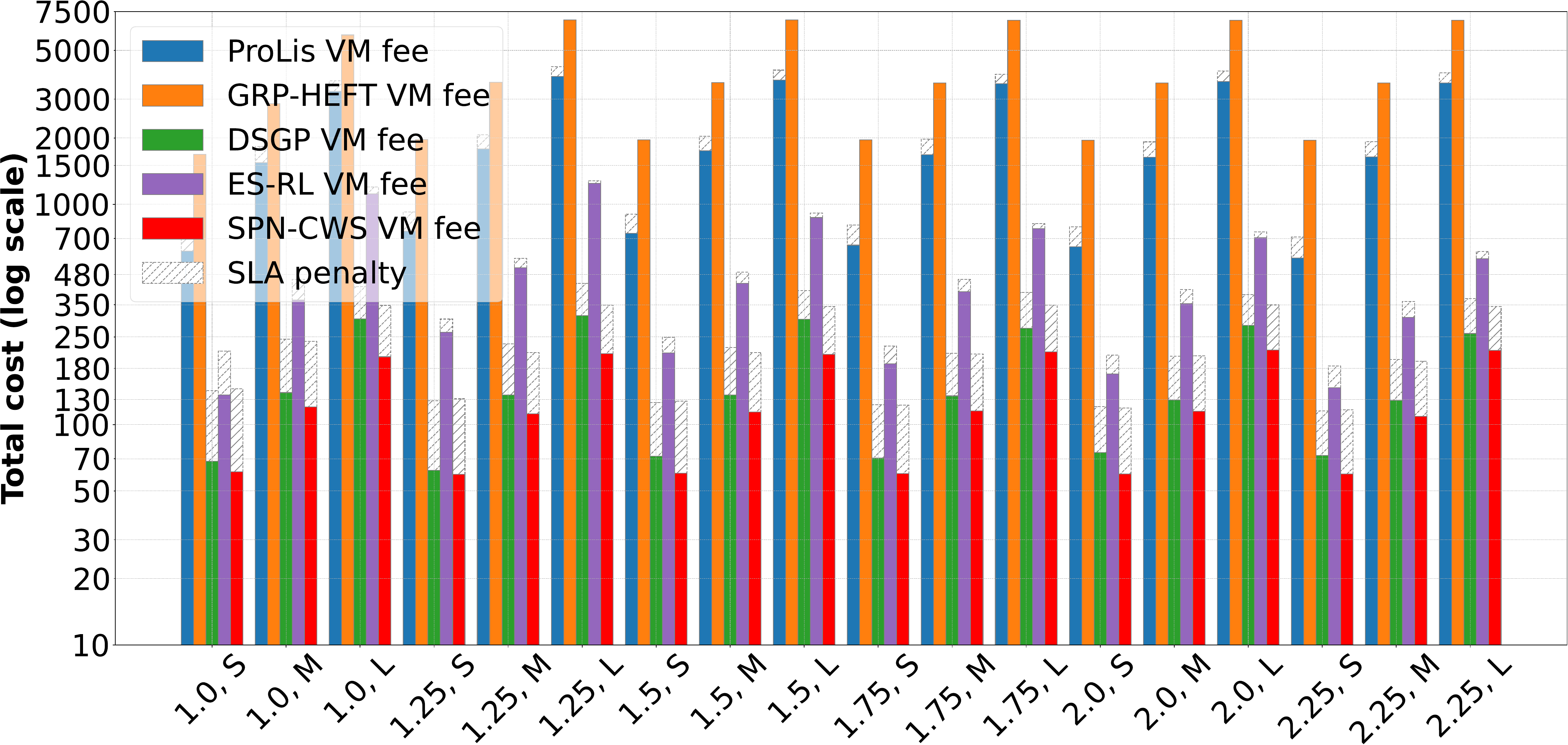} 
\vspace{-2mm}
\caption{The average of VM fees and SLA penalties of all algorithms.}
\label{fig:VM fee and SLA}
\vspace{-8mm}
\end{figure}

Fig. \ref{fig:VM fee and SLA} presents the VM fees and SLA penalties for each algorithm. We can see that GRP-HEFT tends to treat the SLA deadline as a hard constraint (the SLA penalty is 0), leading to excessively high VM rental fees. Similarly, ProLis, DSGP, and ES-RL also tend to produce excessively high VM rental fees. In contrast, SPN-CWS can better balance the trade-off between VM rental fees and SLA penalties, thanks to its capability of using global information to select suitable VM instances. SPN-CWS incurs higher SLA penalties compared to DSGP and ES-RL, but it can significantly reduce the total costs by using cheap VMs, especially on medium and large CDMWS problem scenarios.

\subsection{Convergence Analysis} \label{subsection:Convergence analysis}
\begin{figure}[htbp]
\vspace{-3mm}
\centering
\small
\subfigure[]{\includegraphics[width=.45\columnwidth, height=0.3\hsize]{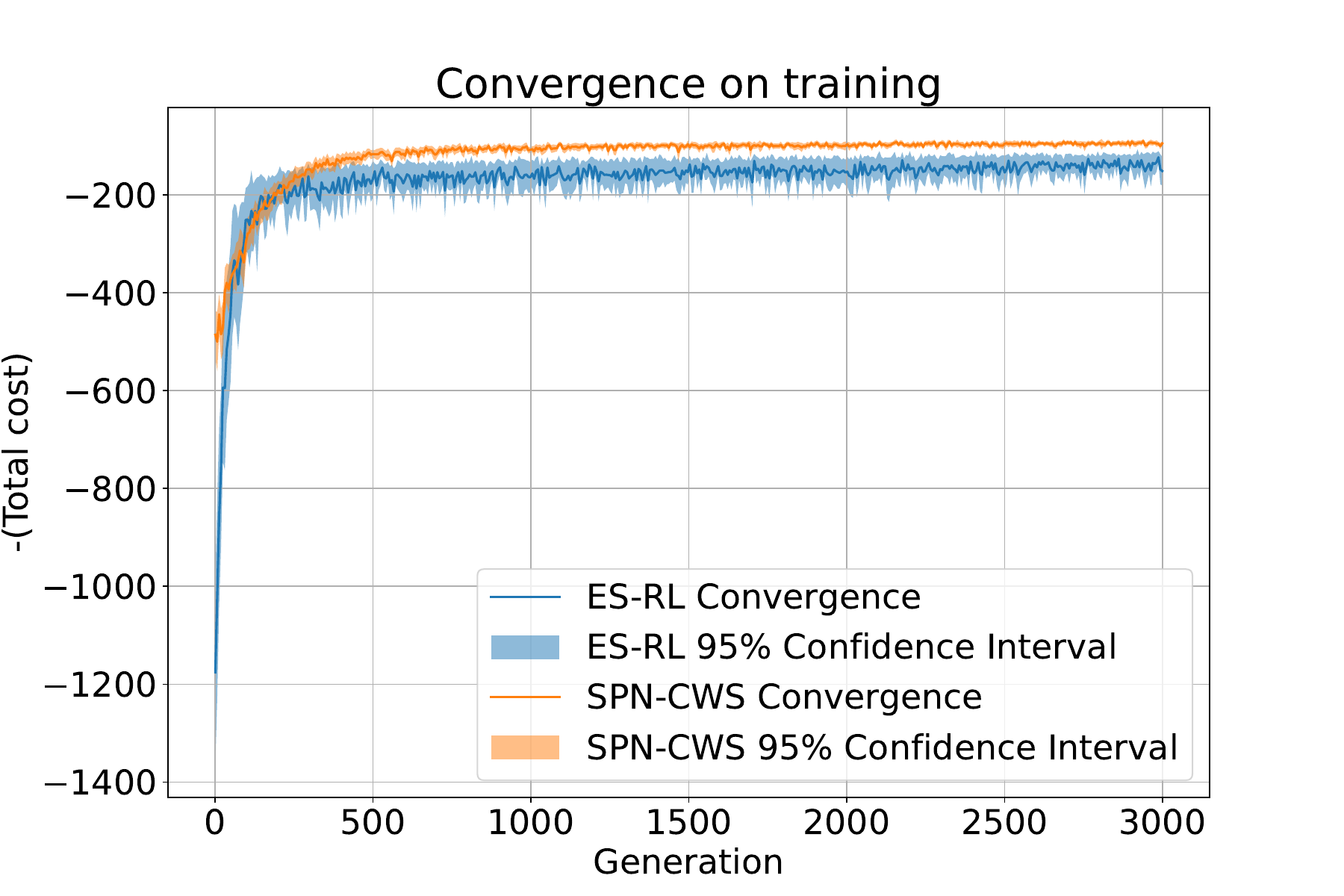}}
\subfigure[]{\includegraphics[width=.45\columnwidth, height=0.3\hsize]{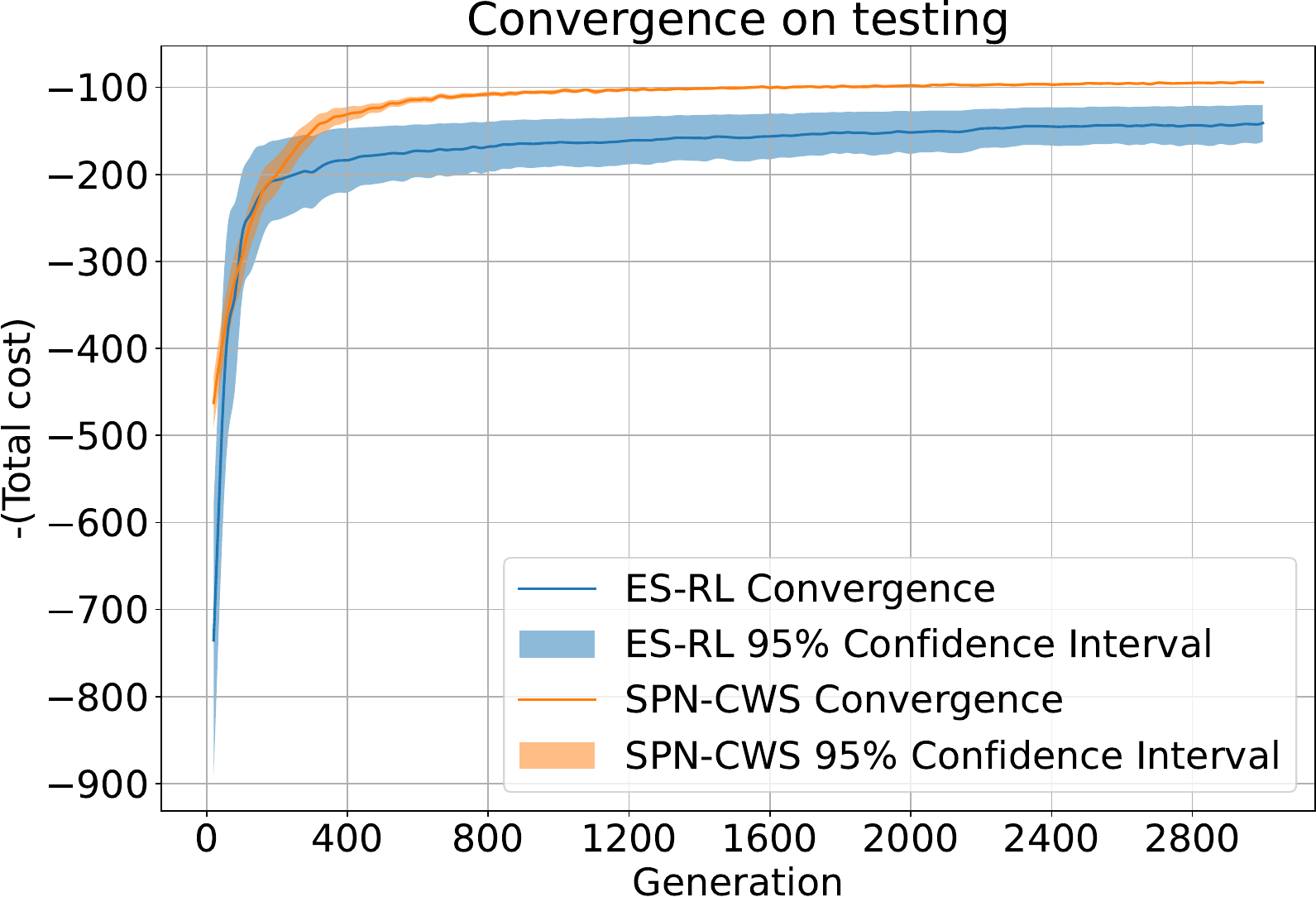}}
\vspace{-5mm}
\caption{The convergence on small-scenario CDMWS instance with $\gamma=5$.}
\label{fig:difference in train and test}
\end{figure}

\vspace{-0.2mm}
Since both ES-RL and SPN-CWS are RL-based algorithms, we compare the convergence behaviors of ES-RL and SPN-CWS during training and testing in Fig. \ref{fig:difference in train and test}. As demonstrated in this figure, SPN-CWS achieves lower total costs compared to ES-RL during training and testing. Notably, despite of using complex network architectures, SPN-CWS achieved competitive convergence speed as that of ES-RL. It also showed faster convergence speed than ES-RL on the testing problems. Furthermore, compared to ES-RL, SPN-CWS enjoys clearly smaller confidence intervals during both training and testing, suggesting that the trained SPN-CWS can perform more consistently and reliably.

\section{Conclusions} \label{section:Conclusions}
This paper introduced a Self-Attention Policy Network (SPN-CWS) for cloud workflow scheduling, capable of capturing global information across all VMs. We developed an ERL system with a cloud simulator to train SPN-CWS efficiently. The trained SPN-CWS selects the most suitable VM for each workflow task by processing all candidate VMs as input. Experimental results demonstrated that our approach significantly outperforms state-of-the-art algorithms in CDMWS and exhibits good convergence speed and stability. Future work will explore online reinforcement learning to enhance adaptability to dynamic workflow changes.

\bibliographystyle{splncs04}
\bibliography{bibliography}

\end{document}